\documentclass[10pt, a4paper]{article}
\usepackage{lrec}
\usepackage{graphicx}
\usepackage{tabularx}
\usepackage{soul}
\usepackage{amsmath}

\usepackage{epstopdf}
\usepackage[utf8]{inputenc}

\usepackage{makecell}
\usepackage{hyperref}
\usepackage{xstring}

\DeclareMathOperator{\embedding}{embedding}
\DeclareMathOperator{\lstm}{LSTM}
\DeclareMathOperator{\concat}{concat}
\DeclareMathOperator{\softmax}{softmax}

\title{Semantic Frame Parsing for Information Extraction : the CALOR corpus\\}

\name{Gabriel Marzinotto$^{1, 2}$, Jeremy Auguste$^2$, Frederic Bechet$^2$, Geraldine Damnati$^1$, Alexis Nasr$^2$}

\address{(1) Orange Labs, Lannion, France\\
	(2) Aix Marseille Univ, Université de Toulon, CNRS, LIS, Marseille, France \\
         \{gabriel.marzinotto, geraldine.damnati\}@orange.com\\
         \{jeremy.auguste, frederic.bechet, alexis.nasr\}@lis-lab.fr\\}

\abstract{
This paper presents a publicly available corpus of French encyclopedic history texts annotated according to the Berkeley FrameNet formalism.
The main difference in our approach compared to previous works on semantic parsing with FrameNet is that we are not interested here in \textit{full text parsing} but rather on \textit{partial parsing}. The goal is to select from the FrameNet resources the minimal set of frames that are going to be useful for the applicative framework targeted, in our case Information Extraction from encyclopedic documents.
Such an approach leverages the manual annotation of larger corpora than those obtained through full text parsing and therefore opens the door to alternative methods for Frame parsing than those used so far on the FrameNet 1.5 benchmark corpus.
The approaches compared in this study rely on an integrated sequence labeling model which jointly optimizes frame identification and semantic role segmentation and identification.
The models compared are CRFs and multitasks bi-LSTMs.
\newline  }

\begin{document}

\maketitleabstract

\section{Introduction}

Semantic Frame parsing is a Natural Language Understanding task that involves detecting in a sentence an event or a scenario, called \textit{Frame}, as well as all the elements or roles that can be associated to this event in the sentence, called \textit{Frame Elements}. One of the most popular semantic frame model is the Berkeley FrameNet project developed by ICSI Berkeley \cite{Baker:1998:BFP:980845.980860}. This model is composed of an inventory of Frames with, for each of them, a list of words, called \textit{Lexical Units} (or LU), that can trigger a frame in a sentence. Besides, for each frame, a list of \textit{Frame Elements} (FE), core or optional, is defined. LUs are pairings of a word with a sense; Frame Elements are the components of a frame, represented by sequences of words in a sentence.


Two kinds of parsing can be done with a Semantic Frame model: \textit{full text parsing} where each word in a sentence is analyzed to check if it can trigger a frame; and \textit{partial parsing} where only a subset of frames and LUs is considered, according to their relevance for a given applicative framework.
Annotating all the possible LUs and frames in a sentence is a very difficult (and expensive) task for human annotators, therefore there are very few corpora annotated this way. Moreover not many languages have such resources.
Most of previous work in semantic frame parsing that has been done with a \textit{full text parsing} approach have used the benchmark corpus FrameNet 1.5. Although the size of this benchmark is relatively large, a lot of frames have a very small number of occurrences in the corpus. This is due to the very large number of frames considered in the semantic model and this makes this corpus particularly challenging for machine learning methods.

On the contrary \textit{partial parsing} can be made on large corpora at a reasonable cost: because the amount of frames and LUs is limited, the annotators can focus only on a few words for each sentence, making the task much easier than full parsing. Corpus obtained this way contain much more examples for each frame, opening the door to more machine learning methods than it is the case with full parsing.

From an applicative point of view, \textit{partial parsing} is also a more realistic option. Although the different senses from the FrameNet model are generic, models trained on the FrameNet 1.5. corpus are not. Assuming that a full text annotation will be available for each new applicative domain is not an option. Moreover a lot of applicative frameworks using semantic models such as frames, like the Information Extraction framework considered in this study, are not interested in full parses but on the contrary only in some specific senses related to the domain targeted.

\begin{figure*}[htbp] 
  \includegraphics[width=1\linewidth]{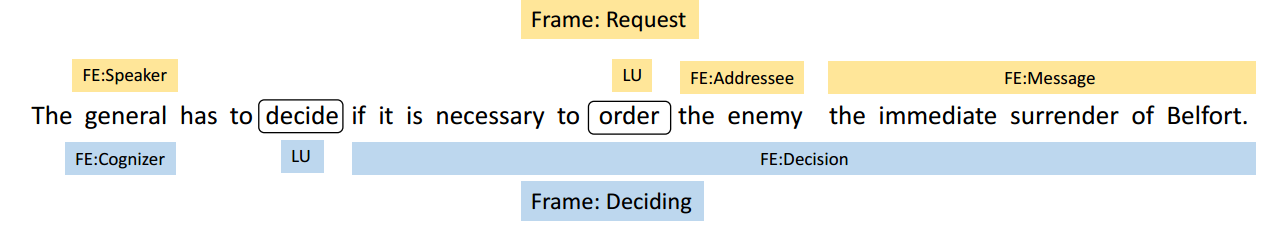}
  \caption{Example of Frame multilabel annotation}
  \label{fig:exemple_frame}
\end{figure*}

An example of this kind of annotation scheme is given in figure \ref{fig:exemple_frame}. As we can see, only two words as considered as lexical units in the sentence: \texttt{decide}, which triggers the frame \texttt{Deciding} and \texttt{order} triggering the frame \texttt{Request}.


This paper presents a study on the use of Sequence Labeling models such as CRF and LSTM for Semantic Frame parsing. Unlike previous studies developing a multi-step approach involving first the Frame identification task, then the frame element detection and labeling tasks, we propose an approach detecting simultaneously LUs and Frame Elements making Frame identification and argument selection an integrated process. A simple heuristic filter is used in order to maintain coherence in the hypotheses produced. We compare two popular sequence labeling methods: conditional random fields and recurrent deep neural networks using Long Short-term Memory.


\section{The CALOR-Frame corpus}

We introduce in this paper the corpus CALOR, which is a collection of documents in French language that were hand annotated in frame semantics. This 1.3M words corpus contains documents from 4 different sources: Wikipedia's Archeology portal (WA, 201 documents), Wikipedia's World War 1 portal (\textit{WGM}, 335 documents), Vikidia's portals of Prehistory and Antiquity (\textit{VKH}, 183 documents) and ClioTexte's~\footnote{https://clio-texte.clionautes.org/} resources about World War One (WW1) (\textit{CTGM}, 16 documents). These sources were chosen to guarantee both writing style and domain diversity. By having documents from Vikidia (an encyclopedia addressed to children from 8-13 years old) and Wikipedia, presenting subjects of ancient history and archeology we can compare the influence of the writing style on the complexity of the task. The same analysis is possible on the WW1 documents, as Cliotexte (a collection of historical documents such as letters, essays, speeches from WW1) and Wikipedia share a common domain with a completely different writing style. This document selection allows to study the importance of the nature of the training data on the performance of the system on a test set on the same subject. Moreover, having data from two different portals of Wikipedia allows to study the domain dependency problem. For example: evaluate if a model for a frame F, trained on data from the archeology domain can successfully be applied on data from the WW1.

In contrast to full text parsing corpus, the frame semantic annotations of CALOR are limited to a small subset of frames from FrameNet \cite{Baker:1998:BFP:980845.980860}.
As described in the introduction, the goal of this \textit{partial parsing} process is to obtain, at a relatively low cost, a large corpus annotated with frames corresponding to a given applicative context.
In our case this applicative context is Information Extraction (IE) from encyclopedic texts, mainly historical texts.

To this purpose we extracted from our 1.3M word text corpus the top 100 most frequent verbs, then we kept those that were more likely to correspond to an action or a situation that would be relevant in our IE context. For example, verbs such as \textit{discover} or \textit{build} are very relevant for exploring archaeological documents.
We looked for the corresponding frames of the selected verbs in the Berkeley FrameNet lexicon and kept a set of 53 different frames.
If a verb could trigger several frames, we only kept those which were relevant in our corpus.

By adding noun triggers to the list of selected verb triggers we obtain a list of 145 Lexical Units (LU), with 30,950 occurrences in the training corpus. 
Selecting the most frequent verbs and nouns from the 1.3M words CALOR corpus as frame triggers is a guarantee that the average number of occurrences per frame is high.
Therefore, even if the list of frames annotated in the CALOR corpus is small compared to the Framenet corpus, we have a large variety of occurrences for each of them, allowing us to build robust parsers for encyclopedic texts, which is the goal of the CALOR corpus.
The list of Frames in CALOR is provided in Table \ref{tabl:FRAMES}.


\begin{table*}
\begin{center}
\scriptsize
\begin{tabular}{ccccc}
Accomplishment & Activity-start & Age & Appointing & Arrest \\
Arriving & Assistance & Attack & Awareness & Becoming \\
Becoming-aware & Buildings & Change-of-leadership & Choosing & Colonization \\
Coming-to-believe & Coming-up-with & Conduct & Contacting & Creating \\
Death & Deciding & Departing & Dimension & Education-teaching \\
Existence & Expressing-publicly & Finish-competition & Giving & Hiding-objects \\
Hostile-encounter & Hunting & Inclusion & Ingestion & Installing \\
Killing & Leadership & Locating & Losing & Making-arrangements \\
Motion & Objective-influence & Origin & Participation & Request \\
Scrutiny & Seeking & Sending & Shoot-projectiles & Statement \\
Subjective-influence & Using & Verification &  & \\
\end{tabular}
\normalsize
\end{center}
\caption{List of Frames annotated in the CALOR corpus}
\label{tabl:FRAMES}
\end{table*}

\subsection{The annotation process}

Once the corpus, the lexical units and the frame set were chosen, we developed an iterative process for the manual annotation of the CALOR corpus.
Preliminarily to this annotation process, the documents were automatically processed by the Macaon~\cite{macaon:2011} tool suite (sentence segmentation, tokenization, POS Tagging, lemmatization, and dependency parsing). Every word within the documents which lemma belongs to the set of selected LUs generates an example to be annotated. It is possible that one sentence generate several examples to annotate if it contains several LUs.
We obtained a set of 30,950 examples to annotate, corresponding to all the LUs occurrences in the CALOR corpus.
Three annotators were hired for this project. Their goal was to process these 30,950 examples: decide for each of them if its corresponding LU triggers or not one of the 53 frames selected, and finally annotate, if a frame was triggered, all its Frame Elements (FE) occurring in the sentence.

In order to reduce the manual annotation time and perform quality control on the corpus produced, we designed an iterative process based on three principles:
\begin{itemize}
\item  an automatic pre-annotation scheme based on the frame parser that will be presented in section \ref{sec:crf};
\item a batch selection process that selects from the unlabeled corpus a set of examples to annotate that  corresponds to the same LU in very similar syntactic and lexical contexts.
\item an automatic quality control estimator that regularly retrains the frame parser and evaluate its performance thanks to a \textit{k-fold} experiments on the part of the corpus already manually annotated.
\end{itemize}

The iterative process based on these principles can be implemented as follows:
\begin{enumerate}
\item \textit{Frame pre-annotation parsing} : an automatic frame parsing process is applied to each example to annotate. It predicts the frame label and the possible FEs for the LU contained in the example. All these automatic annotations are manually checked and eventually corrected by our annotators.
At each iteration the frame parser (see section \ref{sec:crf}) is trained on the subset of the CALOR corpus that is already annotated. For the first iteration, since there is no data to train the parser, all the LUs are labeled with a "\textit{no frame}" label. Each iteration brings more data to train the frame parser.
\item \textit{Batch selection process}: the examples that have not yet been manually processed are sorted according to their LUs, then, using a similarity measure taking into account the lexical and syntactic context in which the LUs occur. They are then grouped into batches that will be sent to the annotators for manual validation.
the goal here is to reduce the cognitive load of the annotators by splitting the corpus to annotate into small batches sharing very similar properties, likely to be annotated the same way.
\item \textit{Manual correction} : a GUI allows annotators to work on the batches of examples produced in the previous step. Annotation is done on text only, no syntactic annotation is provided to the annotators.
The frame pre-annotations produced by the frame parser are displayed, annotators can correct them and add what is missing.
\item \textit{Model training and quality control validation} : at each iteration, the frame parsing models are re-trained on the corpus of manually processed examples. A \textit{k-fold} evaluation is also performed to monitor the evolution of the parsing performance of the model when more validated data is added to the training corpus.
If the frame parsing performance improves, it is a good indication that the added data is coherent with the annotations already processed in the previous iterations.
This can be seen as a quality control measure of the annotation process on the whole dataset, in addition to inter-annotators agreement measures than can also be estimated on small subsets of the corpus.
\end{enumerate}



\subsection{Corpus Statistics}

Table \ref{tab:corpus} presents the distribution of the CALOR corpus among its different sources. We observe that the two first rows of this table, corresponding to Wikipedia, represent most of the corpus.
After the annotation process, 30,950 LU have been annotated, leading to 26,725 LU associated to a frame and 4,225 (13\%) labeled as \textit{OTHER}.
In table \ref{tab:corpus} the columns \textbf{\# Sentences}, \textbf{\# Words}, \textbf{\# Frames}, \textbf{\# Other} and \textbf{\# FE} display the number of sentences, words, frames, LUs, and Frame Elements. \textbf{\% Sentences with Frame} displays the percentage of sentences with at least one frame and \textbf{Lexicon} corresponds to the size of the vocabulary of each document source. The CALOR corpus contains 57,688 FE annotations which averages to 2.2 FE per Frame occurrence.

\begin{table*}
\small
\begin{tabular}{|l|c|c|c|c|c|c|c|}
\hline
\textbf{Document Source}              & \textbf{\# Sentences} & \textbf{\# Words} & \textbf{\# Frames} & \textbf{\# FE} & \textbf{Lexicon} & \textbf{\% Sentence with Frame}  \\ \hline
WGM (Wikipedia WW1)                   & 30994               & 686355            & 14227              & 32708          & 42635  & 34.2\% \\ \hline
WA (Wikipedia Archeology)             & 27023               & 540653            & 9943               & 19892          & 41418  & 28.0\% \\ \hline
CTGM (Cliotexte WW1)                  & 3523                & 67736             & 938                &  1842          & 10844  & 21.4\% \\ \hline
VKH (Vikidia Prehistory \& Antiquity) & 5841                & 85034             & 1617               &  3246          & 11649  & 21.9\% \\ \hline
\hline
\textbf{All}                          & 67381               & 1379778           & 26725              & 57688          & 72127  & 30.0\% \\ \hline
\end{tabular}
\normalsize
\caption{Description of the CALOR corpus}
\label{tab:corpus}
\end{table*}

\begin{figure*}[htbp] 
\begin{center}
\includegraphics[width=0.8\textwidth]{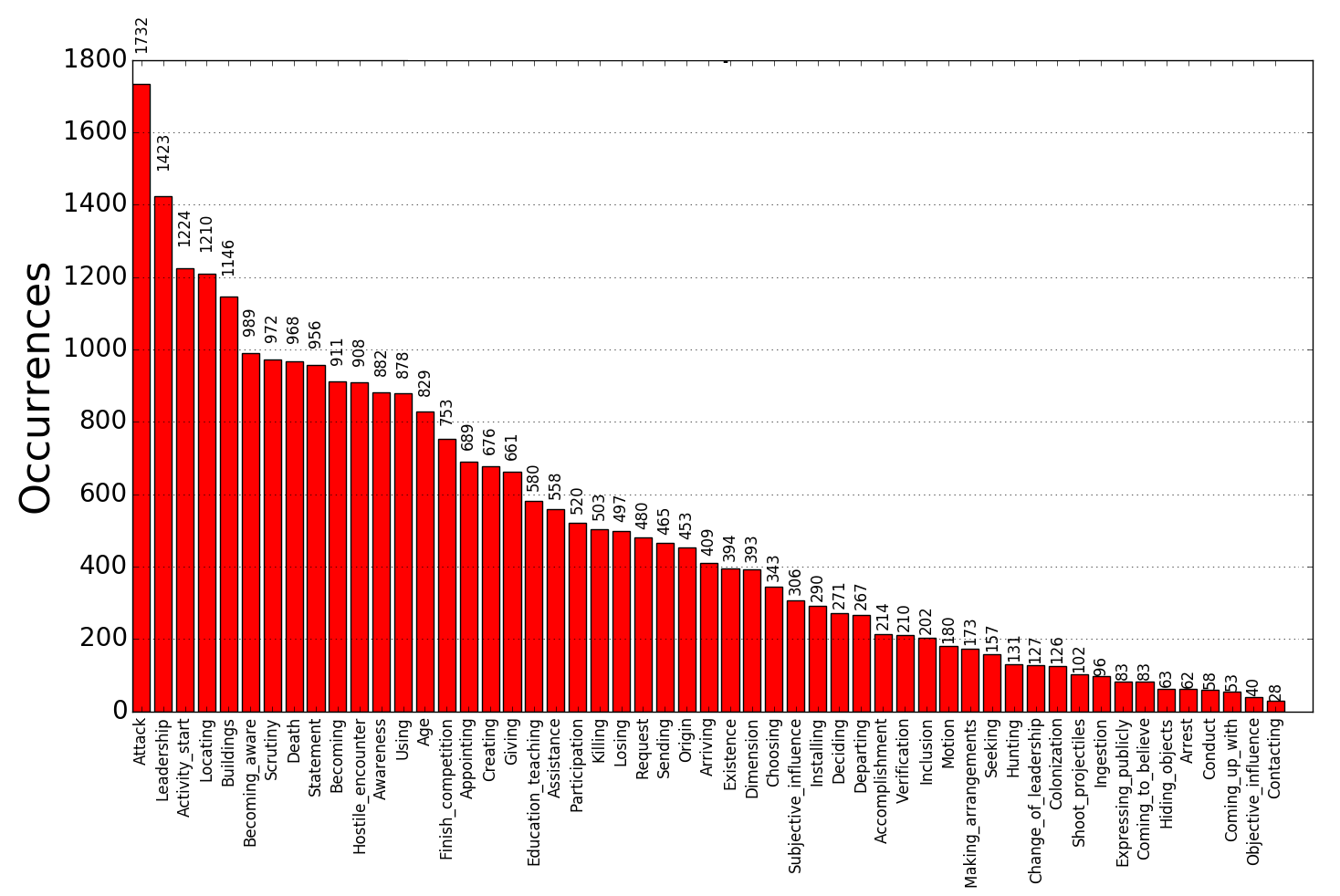}
\end{center} 
\caption{Distribution of the frame occurrences in the CALOR-Frame corpus} 
\label{distrib}
\end{figure*}

Figure \ref{distrib} displays the distribution of the number of annotated examples for each Frame in the CALOR corpus. We observe that the corpus has a large number of examples per frame: half of the frames in CALOR have more than 400 annotated examples and the 10 most frequent frames have more than 900 examples. The most common Frames are \texttt{Attack} (triggers: attaquer, attaque, offensive, bombardement, contre-attaque), \texttt{Leadership} (triggers: commander, diriger, commandement),  \texttt{Activity Start} (triggers: commencer, débuter, commencement, début), \texttt{Locating} (triggers: retrouver, trouver,  localisation) and \texttt{Building} (triggers: construire, fabriquer, élever, construction, fabrication).

\subsection{ Comparison with other corpora }

The comparison between the CALOR-Frame corpus and other corpora with semantic frame annotations is given in table \ref{tab:corpus_compare}.
The column \textbf{Documents} shows the main sources of documents annotated, \textbf{\# Sent} counts the number of sentences in each corpus; \textbf{\% Sent. w/Frame} shows the percentage of sentences that have a Frame annotation; \textbf{Word Lexicon} displays the size of the lexicon of each corpus; \textbf{Frame lexicon}, \textbf{LU lexicon} and \textbf{FE lexicon} correspond to the number of frames, LUs and FEs considered in the annotation model; finally,  \textbf{\# Frame occurrences } shows the number of Frames annotated in the whole corpus.

As we can see, the CALOR-Frame corpus is the only corpus that is not oriented towards journalism, news or current events, it is also the corpus with the largest lexicon size. When we compare it to the existing dataset for Semantic Frame Parsing in  English (SemEval07) and French (ASFALDA)~\cite{candito-etal-2014} we observe that the CALOR corpus has the smallest frame lexicon, but the biggest number of annotations.


\begin{table*}
\small
\begin{tabular}{|l|c|c|c|c|c|c|c|c|c|}
\hline
\textbf{\thead{Corpus \\ Name}} & \textbf{Annotation} & \textbf{Document types} & \textbf{\# Sent} & \textbf{\thead{\% Sent\\ w/ Frame}} & \textbf{\thead{Word\\Lexicon}}  & \textbf{\thead{Frame\\lexicon}} & \textbf{\thead{LU\\lexicon}} & \textbf{\thead{FE\\lexicon}} & \textbf{\thead{\# Frame\\occurrences}}  \\ \hline
SemEval07 & \makecell{English\\ Frames} & Journals & 5946  &  \textbf{85.1\%}  &  14150  & \textbf{720}     & \textbf{3.197}    & \textbf{754}  & 23814   \\ \hline
ASFALDA & \makecell{French\\ Frames} & \makecell{Journals\\ (Le Monde)} & 21634  & 60.7\% &  33955  & 121     & 782      & 140  & 16167  \\ \hline
CALOR & \makecell{French\\ Frames/SRL} & \makecell{Wikipedia\\ Vikidia\\ ClioTexte} & \textbf{67283} &  33.5\% &  \textbf{72127}  & 53      & 145      & 148  & \textbf{26725}   \\ \hline
\end{tabular}
\normalsize
\caption{Semantic Frame corpus comparison}
\label{tab:corpus_compare}
\end{table*}

\section{Frame parsing as a sequence labeling task}
\label{sec:frame_sequence}

As mentioned in the introduction, semantic frame parsing is a structured prediction task where a word can belong to several structures.
For example, in figure \ref{fig:exemple_frame}, the word \texttt{general} can belong to both the frame \texttt{Request} as the frame element \texttt{Speaker} and the frame \texttt{Deciding} as the \texttt{Cognizer}.

In order to consider the Frame parsing task as a \textit{word sequence labeling} task, we need to flatten the frame structures by labeling each word with both semantic and information structure.
In this study we have decided to use the simple \texttt{B,I,O} encoding for word segments where each word label starts with a \texttt{B} if it starts a segment, with an \texttt{I} if the word is inside a segment and \texttt{O} if it doesn't belong to any segment. Links between segments are represented by word indices.

\begin{table}[h]
\scriptsize
\begin{center}
\begin{tabular}{|c|c|c|c|}
\hline
1 & The & \texttt{B:Req:Speaker:11} & \texttt{B:Dec:Cogn:5} \\ \hline
2 & general & \texttt{I:Req:Speaker:11} & \texttt{I:Dec:Cogn:5} \\ \hline
3 & has & \texttt{O} & \texttt{O} \\ \hline
4 & to & \texttt{O} & \texttt{O}\\ \hline
5 & \textbf{decide} &\texttt{O}& \textbf{\texttt{LU:Deciding}} \\ \hline
6 & if &\texttt{O}& \texttt{B:Dec:Decis:5} \\ \hline
7 & it  &\texttt{O}& \texttt{I:Dec:Decis:5} \\ \hline
8 & is &\texttt{O}& \texttt{I:Dec:Decis:5} \\ \hline
9 & necessary &\texttt{O}& \texttt{I:Dec:Decis:5} \\ \hline
10 & to &\texttt{O}& \texttt{I:Dec:Decis:5} \\ \hline
11 & \textbf{order} & \textbf{\texttt{LU:Request}}& \texttt{I:Dec:Decis:5} \\ \hline
12 & the & \texttt{B:Req:Addres:11} & \texttt{I:Dec:Decis:5} \\ \hline
13 & enemy & \texttt{I:Req:Addres:11} & \texttt{I:Dec:Decis:5} \\ \hline
14 & the & \texttt{B:Req:Message:11} & \texttt{I:Dec:Decis:5} \\ \hline
15 & immediate & \texttt{I:Req:Message:11} & \texttt{I:Dec:Decis:5} \\ \hline
16 & surrender & \texttt{I:Req:Message:11}  & \texttt{I:Dec:Decis:5} \\ \hline
17 & of & \texttt{I:Req:Message:11} & \texttt{I:Dec:Decis:5} \\ \hline
18 & Belfort & \texttt{I:Req:Message:11} & \texttt{I:Dec:Decis:5} \\ \hline
\end{tabular}
\caption{Example of corpus with B,I,O format}
\label{table:bio_corpus}
\end{center}
\end{table}

An example of such a representation for the sentence of Figure \ref{fig:exemple_frame} is given in table~\ref{table:bio_corpus}. As we can see column 3 corresponds to the frame \texttt{Request} and column 4 to \texttt{Deciding}.
The LUs triggering the frames are \texttt{order} at index $11$ for the frame  \texttt{Request} and \texttt{decide} at index $5$ for \texttt{Deciding}.
To each frame element is attached the index of the frame it belongs to through a link to the LU that triggered it.

Using such a representation for sequence labeling models like Conditional Random Fields (CRF) or Recurrent Neural Networks (RNN) models is challenging for two reasons:
\begin{enumerate}
\item Multi-labels: each word can receive more than one label according to the number of frames occurring in a sentence. Each label contains the frame as well as the frame element identifiers, therefore there are too many labels to consider building complex labels combining all of them.
\item Linking: an explicit link to the LU triggering a frame is added to each word label of its FEs, as can be seen in table~\ref{table:bio_corpus}. This information is necessary as several expressions of the same frame can occur in a sentence triggered by several LUs.
The absolute value of these links is not meaningful and cannot be predicted the same way as the semantic labels are.
\end{enumerate}

In this study we compare two different strategies in order to deal with the multi-label issue, one based on CRF with a multi-model approach (each LU has its own prediction model) and one based on a bi-LSTM model following a multi-task approach. They are described in the next section.


\section{Sequence labeling models}
  \label{seq:models}

\begin{figure*}[htbp] 
  \includegraphics[width=1\linewidth]{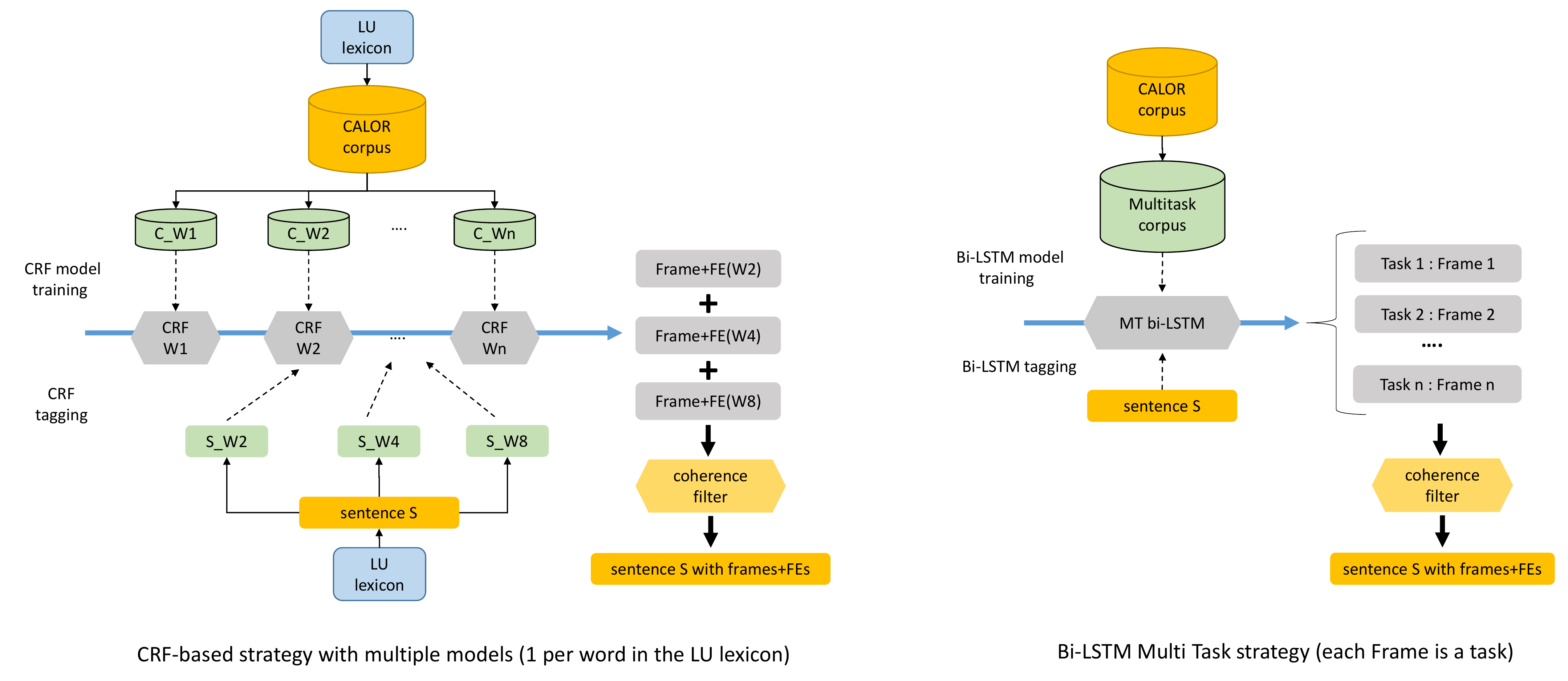}
  \caption{Two different strategies (CRF and bi-LSTM multitask) for Frame parsing}
  \label{fig:models}
\end{figure*}

\subsection{Multi-model CRF approach}
\label{sec:crf}

CRF-based approaches have been used in many NLP tasks involving sequence labeling such as POS tagging, chunking or named entity recognition  \cite{McCallum:2003:ERN:1119176.1119206}.
In order to apply CRF to frame parsing, as described in section~\ref{sec:frame_sequence}, we need to address the multi-label issue.
Since we want to perform frame disambiguation and semantic role detection in one step, and because each word in a sentence cannot trigger more than one frame, we chose a multi-model approach where a CRF-model is trained for each word belonging to the LU lexicon.
This approach is described in Figure~\ref{fig:models}.
At training time, the corpus is split according to the LU lexicon: to each word $W_i$ belonging to this lexicon is attached a sub-corpus containing all the sentences $C_{W_i}$ where $W_i$ occurs.
For each sentence $s\in C_{W_i}$, $W_i$ can trigger a frame $F$ among all the possible frames for this word in the LU lexicon, or nothing.

For example, the sentence shown in table~\ref{table:bio_corpus} will be duplicated into two sub-corpora, $C_{\texttt{order}}$ with column 3 and $C_{\texttt{decide}}$ with column 4.
A CRF model is trained on each $C_{w_i}$ sub-corpus.

At decoding time, when processing a sentence $S$, the same process is applied: first $S$ is duplicated for each word $w_i$ of $S$ belonging to the LU lexicon. Then the CRF model corresponding to each $w_i$ is applied and the different predictions made by the CRF models are merged.

This approach has the advantage of keeping the number of possible labels to predict for each CRF relatively small, limited to the frames that can be triggered by the word considered. Therefore the ambiguity is limited and CRFs can be trained efficiently even with a large number of features.
However the drawback is that the training data is split across words in the LU lexicon, therefore similarities among LU are not exploited.
This situation is acceptable if enough training examples are provided for each LUs, which is the case for the CALOR corpus.

\subsection{Multi-task LSTM approach}

Deep Neural Networks (DNN) with word embedding is the state of the art approach for semantic frame parsing~\cite{hermann2014semantic}.
More recently \textit{recurrent neural networks} (RNN) with Long Short Memory (LSTM) cells have been applied to several semantic tagging tasks such as \textit{slot filling} ~\cite{mesnil2015using} or even frame parsing ~\cite{hakkani2016multi,tafforeau2016joint} for Spoken Language Understanding.

Following these previous works, we propose in this study a single-layered bidirectional LSTM sequence to sequence architecture to perform frame tagging.
To deal with the multi-label issue we could train a biLSTM model per LU, using the same approach as for the CRF, however, the number of examples per LU is reduced and neural networks do not perform well on small datasets. We would face the same problem if instead we decided to train one biLSTM per frame.
The remaining possibility is to build a single biLSTM to predict all possible frames in order to automatically learn a feature representation meaningful to all frames.
We chose a multi-task approach (biLSTM-MT) similar to the one proposed by \cite{tafforeau2016joint}, which models each frame as an isolated task. 
In this model two LSTM models, one forward and one backward are concatenated and shared among all tasks.
Then a task-specific fully-connected output layer is added for each task.

In this work we consider each frame of our FrameNet model as a different task.
This approach is described in figure~\ref{fig:models}.
At decoding time each sentence is processed by the network and a distribution probability on the labels of each task for each word is produced.
We only keep the labels above a certain threshold.

\subsection{Coherence filter}
Once the sequence tagging process is performed, each word is labeled with a frame element and position labels or a null label ($O$) as presented in table \ref{table:bio_corpus}. Because the labels given at the word level might not be coherent at the frame level, we apply a \textit{coherence filter} to the output of the tagging process. This filter is in charge of removing incoherences (FEs not starting with a $B$ label; FEs without a frame) and linking the FEs to the LU that triggered the corresponding frame.

This filter implement is a very simple strategy: in a given sentence, if a word $W$ of index $i$ is labeled as a LU trigger for frame $F$, we link all the FEs detected in the sentence with the same frame label $F$ to the LU $w_i$. At the end of this process, all FEs that have not been linked to a LU are removed.




\subsection{Feature selection}


The feature sets used for the CRF and the biLSTM approaches differ.
For the CRF model, each training sample contains only one trigger word which is clearly identified, therefore we can use this information in order to add global constraints on the feature set of each word to process.
On the opposite the multi-task biLSTM models can process several triggers in the same sentence, therefore the feature set cannot be biased toward a specific trigger and only local features are considered.

For the CRF models we consider 3 features: word lemma, part-of-speech (POS) and the syntax dependency path between the word to process and the potential frame trigger in the sentence. This dependency path is built through the concatenation of the syntactic functions between the word and the trigger.
In the general case, a trigger is not necessarily at the root of the syntactic tree, for this reason, the dependency paths are composed of both links from child to parent (ascending links) and from parent to child (descending links) we make distinction of both types of links in the way we encode the dependency path.     

For the biLSTM-MT we consider 4 features: word embeddings (Glove embeddings of 200 dimensions trained on French Wikipedia), POS, syntactic function (without a link) and a boolean indicator of whether the word belongs to the lexicon LU or not. All the features are encoded as trainable embeddings, we allow the network to adapt them during the training of the frame parsing task.  

For both systems, in order to extract lemmas, POS and syntactic dependency trees we processed the frame annotated corpus using  MACAON \cite{macaon:2010} trained with a set of POS and dependencies similar to the one proposed in the Paris French TreeBank \cite{abeille2003building,abeille2004enriching}.

The main differences between the multi-model CRF and multi-task biLSTM models are summarized in table \ref{table:LSTM_CRF_Prior_Comparaison}.

\begin{table*}[h]
  \begin{center}
    \begin{tabular}{ c | c | c |  }
      \cline{2-3}  & CRF-LU & biLSTM-MT  \\ \hline
      \multicolumn{1}{ |c|  }{ Lexical Feat }  & Lemma & Glove Word Embeddings   \\ \hline
      \multicolumn{1}{ |c|  }{ Morpho-Syntactic Feat } & POS & POS Embeddings   \\ \hline
      \multicolumn{1}{ |c|  }{ Syntactic Feat }       & Dep. Path to Trigger  & Syntactic Function Embeddings   \\ \hline
      \multicolumn{1}{ |c|  }{ Model }        & 145 CRF (1 CRF per LU) & 1 biLSTM 54 Taks (1 Task per Frame) \\ \hline
      \multicolumn{1}{ |c|  }{ Scaling }      & Many LU & Many Taks   \\ \hline
      \multicolumn{1}{ |c|  }{ Labeling Process }    & 1 Application per LU per Phrase & Parse all the Frames in 1 Application    \\ \hline
       \multicolumn{1}{ |c|  }{ Information Sharing }   & No Sharing & Implicit Sharing Between Frames     \\ \hline
    \end{tabular}
    \caption{ Comparative Overview of CRF-LU and biLSTM-MT models }
    \label{table:LSTM_CRF_Prior_Comparaison}
  \end{center}
\end{table*}

\section{Evaluation}

\subsection{\label{corpus}Experimental setup}

When a sentence is processed, there are 4 steps or sub-tasks that take place either explicitly or implicitly in the frame parsing process.
Even though our approaches perform frame detection and semantic role labeling in one step, we detail the scores for each sub-task because they are relevant indicators of the performance of a parser, and they serve as point of comparison between our models.
These 4 sub-task are:
\begin{enumerate}
\item \textit{trigger identification} (TI) which decides whether a word in a sentence can trigger a frame or not;
\item \textit{trigger classification} (TC) which assigns the frame label to the trigger word detected to form a LU;
\item \textit{role filler identification} (RI) which detects potential semantic role fillers in the sentence for the frame detected;
\item \textit{role filler classification} (RC) which assigns a label to each role filler detected in order to obtain the \textit{Frame Elements} (FE) of the frame detected.
\end{enumerate}
In this study we consider these 4 tasks as a cascade process: an error in task 1 will lead to several errors in task 4 since no correct FEs will be detected if the frame is not triggered.

The CALOR corpus is annotated with a small number of frames compared to the FrameNet 1.5 corpus, therefore the ambiguity for subtasks 1 and 2 is rather small.
The most complex task is of course task 4, \textit{role filler classification}, since every detection and classification of LUs and FEs has to be correct.
That is why we will pay a special attention to this task to compare the models in the following experiments.

To carry out our experiments we divide the corpus CALOR assigning 80\% of the frame occurrences to the train set and 20\% to the test set.
This split is done in such a way that the frame distribution remains as similar as possible between train and test while considering a document as an atomic unit that cannot be subdivided and should be either in the train or in the test set. This split does not take into account the LU distribution, a LU can therefore appear both in train and test, only in train or only in test.

Similarly to previous work on semantic frame parsing we will use \textit{precision}, \textit{recall} and \textit{f-measure} on the 4 subtasks presented in order to evaluate our approaches.
We also compute \textit{precision/recall} curves by using different acceptation thresholds on the frame and semantic role hypotheses output by our models.
In this study we set the operating point for comparing our models to the \textit{Equal Error Rate} (EER) between the precision and recall measures.

\subsection{Overall results}

The comparison between the multi-model CRF (CRF-MM) and multi-task biLSTM (biLSTM-MT) performance over the set of 4 sub-tasks presented in the previous section is presented in Table~\ref{table:Global_Performance}.
Both systems use their full feature set as presented in table \ref{table:LSTM_CRF_Prior_Comparaison} with EER between precision and recall as the chosen operating point.
As expected performance for subtasks 1 and 2 are very good, much better than those reported on the FrameNet 1.5 corpus~\cite{hermann2014semantic}.
This is due to the \textit{partial parsing} approach used here where only a subset of the FrameNet model is used.

CRF-MM model performs better on the trigger identification and classification tasks (subtasks 1 and 2) while the biLSTM-MT model is better on role identification and classification (subtasks 3 and 4).
The reason for this behavior is that it is easier for a CRF model to identify the proper frame for each trigger word, as they can only trigger a few frames, while for the biLSTM-MT all frames are in competition.
This reduction of ambiguity for CRF-MM models has a cost: splitting the training corpus according to the LU lexicon and therefore reducing the training data for each model.
If this is not an issue for the two low ambiguity subtasks 1 and 2, the situation is different for subtasks 3 and 4.
Indeed biLSTM-MT is a single model that learns from all training data, it is able to automatically learn relevant features and capture semantic aspects of text using a shared layer and then use these features to classify tokens in roles of different frames learned one by tasks.
This ability to use the whole corpus leads the biLSTM-MT model to outperform CRF-MM for tasks 3 and 4.

\begin{figure}[htbp] 
  \includegraphics[width=1\linewidth]{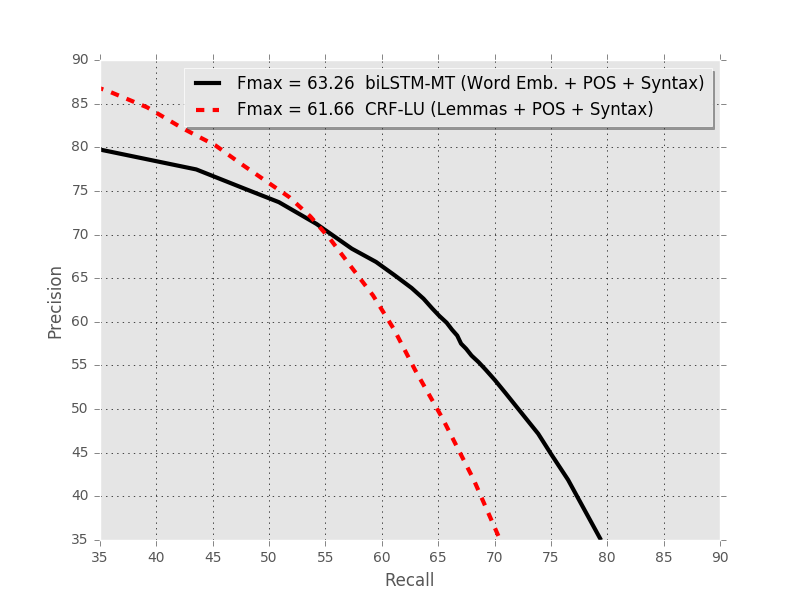}
  \caption{Precision Recall curves for CRL-LU and biLSTM-MT on the full feature set }
  \label{fig:parameter_influence_bests}
\end{figure}

This result is confirmed by figure \ref{fig:parameter_influence_bests} which displays the precision/recall curves of both methods on subtask 4.
As we can see, biLSTM-MT is better in terms of maximal F-measure.
It is interesting to observe that both models do not show the same precision-recall trade-off.
The CRF-MM is able to parse text at a high precision with a low recall, this is not the case of biLSTM-MT, that handles a much larger set of frames and is more prone to precision errors.
On the other hand, biLSTM-MT achieves a better recall at a fairly good precision, this happens because it is trained on the full dataset and it is able to learn syntactic patterns from different frames and extrapolate them from one frame to the other. 


\begin{table}[h]
  \begin{center}
    \begin{tabular}{ c | c | c | }
      \cline{2-3}
       &  \multicolumn{1}{|p{1.4cm}|}{\centering Fmeasure \\ CRF-MM} & \multicolumn{1}{|p{1.4cm}|}{\centering Fmeasure \\ biLSTM-MT} \\ \hline
      \multicolumn{1}{ |c|  }{ Trigger Ident. (TI)} & $ 96.4 $ & $ 95.5 $ \\ \hline
      \multicolumn{1}{ |c|  }{ Trigger Classif. (TC)} & $ 95.2 $ & $ 92.3 $ \\ \hline
      \multicolumn{1}{ |c|  }{ Role Ident. (RI)}    & $ 69.7 $ & $ 70.3 $ \\ \hline
      \multicolumn{1}{ |c|  }{ Role Classif. (RC)}    & $ 60.6 $ & $ 63.2 $ \\
      \hline
    \end{tabular}
    \caption{Comparison of our two models on their full feature set operating at EER  }
    \label{table:Global_Performance}
  \end{center}
\end{table}

\section{Conclusion}
In this paper we introduced the CALOR corpus as well as a comparison between two different semantic frame parsing models that consider the task as a sequence labeling task.
The main contribution of this work is to propose a new corpus, publicly available, that contain semantic frame annotations that differ from previous corpus such as FrameNet and SemEval.
In our case only \textit{partial} annotation is considered, allowing to annotate much larger corpora at a lower cost than full text annotation.
Only a small subset of the FrameNet lexicon is used, however the amount of data annotated for each frame is much larger than in other corpora, allowing to develop and test different parsing methods.

In this study two frame parsing models are compared, one based on CRF following a multi-model approach and other based on biLSTM with a multi-task approach.
Experiments show that biLSTM-MT model achieves a better recall, while CRF-MM achieves better precision, this is due to the architecture of each model,  in CRF-MM we divide frame parsing into small subtasks one per LU, reducing the number of possible labels in each decision, thus augmenting the precision.
On the other hand biLSTM-MT is able to share data across LUs boosting its capacity to deal with complex syntactic patterns and being able to retrieve more frame elements during parsing.
\section{Bibliographical References}
\label{main:ref}

\bibliographystyle{lrec}
\bibliography{xample}


\end{document}